\documentclass{article}
\usepackage{ijcai19}

\usepackage{times}
\usepackage{soul}
\usepackage{url}
\usepackage[utf8]{inputenc}
\usepackage[small]{caption}
\usepackage{graphicx}
\usepackage{amsmath}
\usepackage{booktabs}
\usepackage{algorithm}
\usepackage{algorithmic}
\usepackage[draft]{hyperref}
\usepackage{adjustbox}
\usepackage[none]{hyphenat}

\title{Learning Topic Models by Neighborhood Aggregation}
\author{
	Ryohei Hisano
	\affiliations
	Graduate School of Information Science and Technology, The University of Tokyo, Tokyo, Japan
	\emails
	em072010@yahoo.co.jp
}

\begin{document}

\maketitle

\begin{abstract}
	
Topic models are frequently used in machine learning owing to their high interpretability and modular structure.  However, extending a topic model to include a supervisory signal, to incorporate pre-trained word embedding vectors and to include a nonlinear output function is not an easy task because one has to resort to a highly intricate approximate inference procedure. The present paper shows that topic modeling with pre-trained word embedding vectors can be viewed as implementing a neighborhood aggregation algorithm where messages are passed through a network defined over words. From the network view of topic models, nodes correspond to words in a document and edges correspond to either a relationship describing co-occurring words in a document or a relationship describing the same word in the corpus.  The network view allows us to extend the model to include supervisory signals, incorporate pre-trained word embedding vectors and include a nonlinear output function in a simple manner.  In experiments, we show that our approach outperforms the state-of-the-art supervised Latent Dirichlet Allocation implementation in terms of held-out document classification tasks.

\end{abstract}

\section{Introduction}

Topic models are widely used in both academia and industry owing to their high interpretability and modular structure \cite{Blei2003}.  The highly interpretable nature of topic models allows one to gain important insights from a large collection of documents while the modular structure allows researchers to bias topic models to reflect additional information, such as supervisory signals \cite{Mcauliffe2008}, covariate information \cite{Ramage2009}, time-series information \cite{Park2015} and pre-trained word embedding vectors \cite{Nguyen2015}.

However, inference in a highly structured graphical model is not an easy task.  This hinders practitioners from extending the model to incorporate various information besides text of their own choice. Furthermore, adding a nonlinear output function makes the model even more difficult to train.

The present paper shows that topic modeling with pre-trained word embedding vectors can be viewed as implementing a neighborhood aggregation algorithm \cite{Hamilton2017} where the messages are passed through a network defined over words.  From the network view of topic models, Latent Dirichlet Allocation (LDA) \cite{Blei2003} can be thought of as creating a network where nodes correspond to words in a document and edges corresponds to either a relationship describing co-occurring words in a document or a relationship describing the same word in the corpus.  The network view makes it clear how a topic label configuration of a word in a document is affected by neighboring words defined over the network and adding supervisory signals amounts to adding new edges to the network.  Furthermore, by replacing the message passing operation with a differentiable neural network, as is done for neighborhood aggregation algorithms \cite{Hamilton2017}, we can learn the influence of the pre-trained word embedding vectors to the topic label configurations as well as the effect of the same label relationship, from the supervisory signal.

Our contribution is summarized as follows.

\begin{itemize}
	\item We show that topic modeling with pre-trained word embedding vectors can be viewed as implementing a neighborhood aggregation algorithm where the messages are passed through a network defined over words.
	\item By exploiting the network view of topic models, we propose a supervised topic model that has an adaptive message passing where the parameters governing the message passing and the parameters governing the influence of the pre-trained word embedding vectors to the topic label configuration is learned from the supervisory signal.
	\item Our model includes a nonlinear output function connecting document to their corresponding supervisory signal.
	\item Our approach outperforms state-of-the-art supervised LDA implementation \cite{Katharopoulos2016,Zhu2009} on a wide variety of datasets regarding predictive performance and gives a comparative performance concerning topic coherence.
\end{itemize}

\section{Notations}

We briefly summarize the basic notations used throughout the paper.  Let $1 \leq d \leq D$, $1 \leq w \leq W$, $1 \leq k \leq K$ and $1 \leq s \leq S$ respectively denote the document index, word index, topic number index and label index.  We denote by $x_{d,w}$ the number of word counts for a particular document--word pair $(d,w)$.  The task of topic modeling is to assign the average topic label configuration $z={z_{d,w}^{k}}$ from the observed document word count matrix $X=\{x_{d,w}\}$ where the average topic label configuration is defined as  $z_{d,w}^{k} := \Sigma_{i=1}^{x_{d,w}} \frac{z_{d,w,i}^{k}}{x_{d,w}}$ for all document--word pairs $(d,w)$ in the corpus.  The average topic label configuration sums to 1 over topic index $1 \leq k \leq K$ (i.e., $\Sigma_{k=1}^{K} z_{d,w}^{k}=1)$ and one of the tasks in this paper is to calculate the messages (which we denote $\mu_{d,w}^{k}$) that can approximate all $z_{d,w}^{k}$ in the corpus.  We denote by $\mu_{d,w}^{k}(t)$ the estimated message at iteration $t$.  Omission of the subscript $k$ (i.e., $\mu_{d,w}(t)$) simply implies the vectorized form of $\mu_{d,w}^{k}(t)$s (i.e., $[\mu_{d,w}^{1}(t) \cdots \mu_{d,w}^{K}(t)]$).

We use $\theta_{d}$ to denote the topic proportion distribution of document $d$ and $\phi_{w}$ to denote the topic distribution.  We also use $v_{i}$ to denote node attribute information attached to node $i$. 

\begin{figure}
	\centering
	\includegraphics[width=0.7\linewidth]{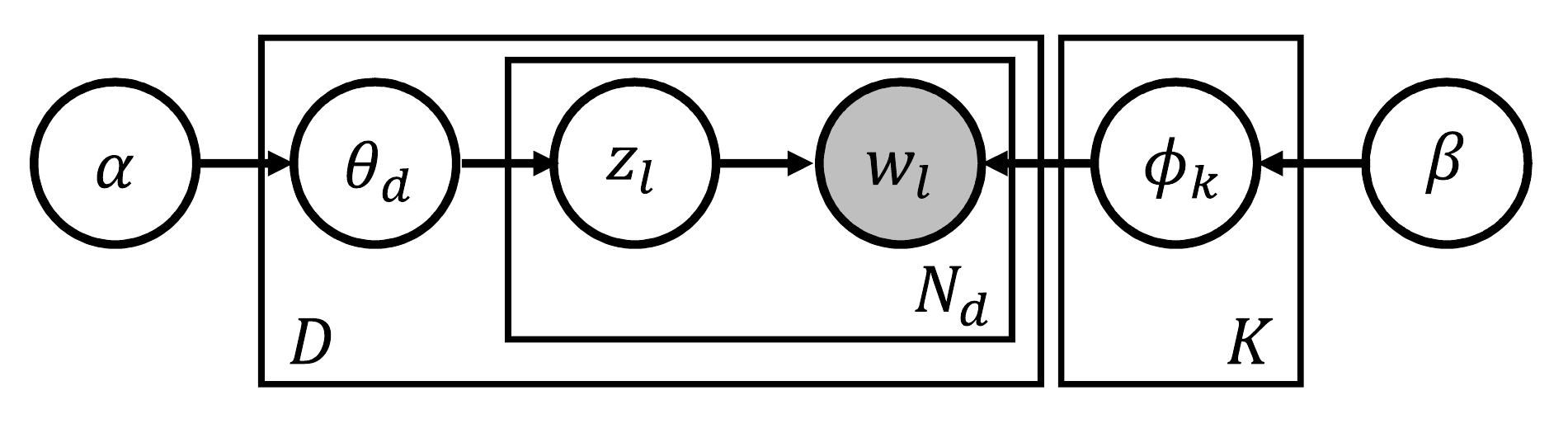}
	\caption{Bayesian network plate diagram representation of LDA}
	\label{figlda}
\end{figure}

\section{Background}
\subsection{Factor Graph Approach to LDA}

From the perspective of topic modeling, our approach is related to the work of \cite{Zeng2013}, who reframed LDA as a factor graph and used the message passing algorithm  for inference and parameter estimation.  

The classical Bayesian network plate diagram representation of LDA is presented in Figure~\ref{figlda}.  The joint distribution of the model can be summarized as 

\begin{equation}
\label{LDA}
\begin{split}
p(x,z|\alpha,\beta) = \prod_{d=1}^{D} \prod_{k=1}^{K} \frac{\Gamma(\Sigma_{w=1}^{W} x_{d,w}z_{d,w}^{k} + \alpha)}{\Gamma(\Sigma_{k=1}^{K}(\Sigma_{w=1}^{W}x_{d,w}z_{d,w}^{k} + \alpha))}\times \\
\prod_{w=1}^{W} \prod_{k=1}^{K} \frac{\Gamma(\Sigma_{d=1}^{D} x_{d,w} z_{d,w}^{k} + \beta)}{\Gamma(\Sigma_{w=1}^{W}(\Sigma_{d=1}^{D} x_{d,w} z_{d,w}^{k} + \beta))},
\end{split}
\end{equation}

\noindent where $\alpha$ and $\beta$ denote hyperparameters of the Dirichlet prior distribution.  Meanwhile, by designing the factor function as

\begin{equation}
\begin{split}
f_{\theta_{d}}(x,z,\alpha) = \prod_{k=1}^{K} \frac{\Gamma(\Sigma_{w=1}^{W} x_{d,w}z_{d,w}^{k} + \alpha)}
{\Gamma(\Sigma_{k=1}^{K}(\Sigma_{w=1}^{W}x_{d,w}z_{d,w}^{k} + \alpha))}
\end{split}
\end{equation}

\noindent and

\begin{equation}
\begin{split}
f_{\phi_{w}}(x,z,\beta) = \prod_{k=1}^{K} \frac{\Gamma(\Sigma_{d=1}^{D} x_{d,w} z_{d,w}^{k} + \beta)}{\Gamma(\Sigma_{w=1}^{W}(\Sigma_{d=1}^{D} x_{d,w} z_{d,w}^{k} + \beta))},
\end{split}
\end{equation}

\begin{figure}
	\centering
	\includegraphics[width=.6\linewidth]{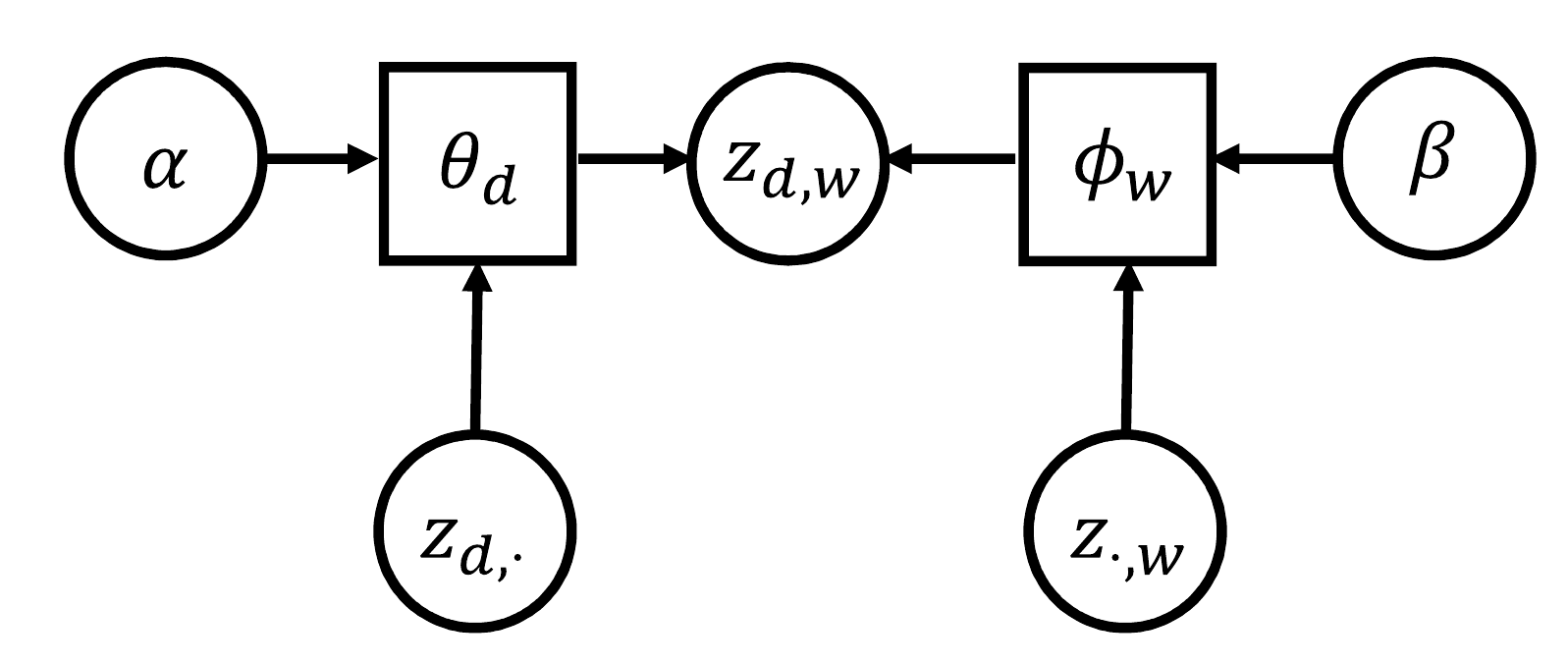}
	\caption{Factor graph representation of LDA}	
	\label{fig:factor}
\end{figure}

\noindent a factor graph representation of LDA (i.e., Figure~ \ref{fig:factor}) can be summarized as 

\begin{equation}
\label{factorgraph}
\begin{split}
p(x,z|\alpha,\beta) = \prod_{d=1}^{D} f_{\theta_{d}}(x,z,\alpha) \prod_{w=1}^{W} f_{\phi_{w}}(x,z,\beta).
\end{split} 
\end{equation}

\noindent Equation~(\ref{factorgraph}) shows that the factor graph representation encodes exactly the same information as Eq. (\ref{LDA}).  From the factor graph representation, it is possible to reinterpret LDA using a Markov random field framework and thus infer messages for words in a document using loopy belief propagation. The essence of their paper can be summarized by a message updating equation of the form

\begin{equation}
\label{MesPasLDA}
\begin{split}
\mu_{d,w}^{k}(t+1) \propto \frac{\Sigma_{w'=1,w' \neq w}^{W} x_{d,w'} \mu_{d,w'}^{k}(t) + \alpha}{\Sigma_{k=1}^{K} (\Sigma_{w'=1,w' \neq w}^{W} x_{d,w'} \mu_{d,w'}^{k}(t) + \alpha)} \times \\
\frac{\Sigma_{d'=1, d' \neq d}^{D} x_{d',w} \mu_{d',w}^{k}(t) + \beta}{\Sigma_{w=1}^{W} ( \Sigma_{d'=1, d' \neq d}^{D} x_{d',w} \mu_{d',w}^{k}(t) + \beta)}.
\end{split}
\end{equation}

Our goal in this paper is to connect the factor graph approach of LDA with the neighborhood aggregation algorithm.

\subsection{Neighborhood Aggregation Algorithm}

The goal of node embedding is to represent nodes as low-dimensional vectors summarizing the positions or structural roles of the nodes in a network \cite{Hamilton2017}.  The neighborhood aggregation algorithm is a recently proposed algorithm used in node embedding \cite{Dai2016,Hamilton2017}, which tries to overcome limitations of the more traditional direct encoding \cite{Perozzi2014} approaches.  The heart of a neighborhood aggregation algorithm lies in designing encoders that summarize information gathered by a node's local neighborhood.  In a neighborhood aggregation algorithm, it is easy to incorporate a graph structure into the encoder, leverage node attribute information and add nonlinearity to the model.  The parameters defining the encoder can be learned by minimizing the supervised loss \cite{Dai2016,Hamilton2017}.  In \cite{Dai2016}, it was shown that these algorithms can be seen as replacing message passing operations with differential neural networks.  We refer to \cite{Dai2016} for a more detailed explanation.

The essence of neighborhood aggregation algorithms is characterized by three steps: aggregation, combination and normalization.  In the aggregation step, we gather information from a node's local neighborhood.  A concatenation, sum-based procedure or elementwise mean is usually employed.  In the combination step, we combine a node's attribute information with the gathered information from the aggregation step.  After passing the combined information through a nonlinear transformation, we normalize the message so that the new updated messages can be further used by neighboring nodes.

The overall process is succinctly summarized as

\begin{equation}
\label{NeighborAgg}
\begin{split}
\mu_{i}(t+1) = Norm(\sigma(Comb(v_{i},Agg(\mu_{j}(t);\\
\forall j \in Nei(i));W_{A},W_{C}))),
\end{split}
\end{equation}

\noindent where $\mu_{i}(t+1)$ denotes the message of node $i$ at iteration $t+1$; $Nei(i)$ denotes neighboring nodes of node $i$; $Agg$, $Comb$ and $Norm$ respectively denote aggregation, combination and normalization functions; $W_{A}$ and $W_{C}$ denote parameters used in the combination step (e.g., $Comb(v_{i},x_{i}) = W_{C}v_{i} + W_{A}x_{i}$) and $\sigma$ is an elementwise nonlinear transformation.  With enough update iterations, the model converges and the desired messages are learned.  

Suppose that we are given a training dataset $D = \{X,y_{1:D}\}$, where $X$ is the observed network and $y_{1:D}$ is the supervisory signal attached to each node.  For a classification problem $\{y_{d} \in {1,\cdots,S}\}$, we use a simple neural network with softmax output to transform the learned messages to probabilities and to minimize the cross-entropy loss.  This is summarized as

\begin{equation}
\label{softmax}
\begin{split}
score_{d} = S_{C} \sigma(S_{B} \sigma(S_{A} \mu_{d}  + T_{A}) + T_{B}) + T_{C} \\
p_{d}^{s} = \frac{exp(score_{d}^{s})}{\Sigma_{s=1}^{S} exp(score_{d}^{s})} \\
min_{\{ W_{A},W_{C},S_{A},S_{B} \}} -\Sigma_{d=1}^{D} \Sigma_{s=1}^{S} y_{d}^{s} log(p_{d}^{s})
\end{split}
\end{equation}

\noindent where $S_{A}$, $S_{B}$, $S_{C}$ are either $H_{1} \times K$, $H_{2} \times H_{1}$ and $S \times H_{3}$ weight matrices transforming the messages and $T_{A}$, $T_{B}$ and $T_{C}$ are bias vectors of either size $H_{1}$, $H_{2}$ and $S$, $y_{d}^{s}=1$ if the label of node $d$ is $s$ and zero otherwise and $\sigma$ is an elementwise nonlinear transformation.  We use the classic sigmoid function in this paper.

For a regression problem $y_{n} \in R$, parameters can be learned by minimizing the sum-of-squared loss:

\begin{equation}
\begin{split}
min_{\{W_{A},W_{C},S_{A},S_{B}\}} \Sigma_{d=1}^{D} (y_{d} - score_{d})^{2}.
\end{split}
\end{equation}

\begin{figure*}[ht]
	\centering
	\begin{minipage}{.49\textwidth}
		\centering
		\includegraphics[width=.65\linewidth]{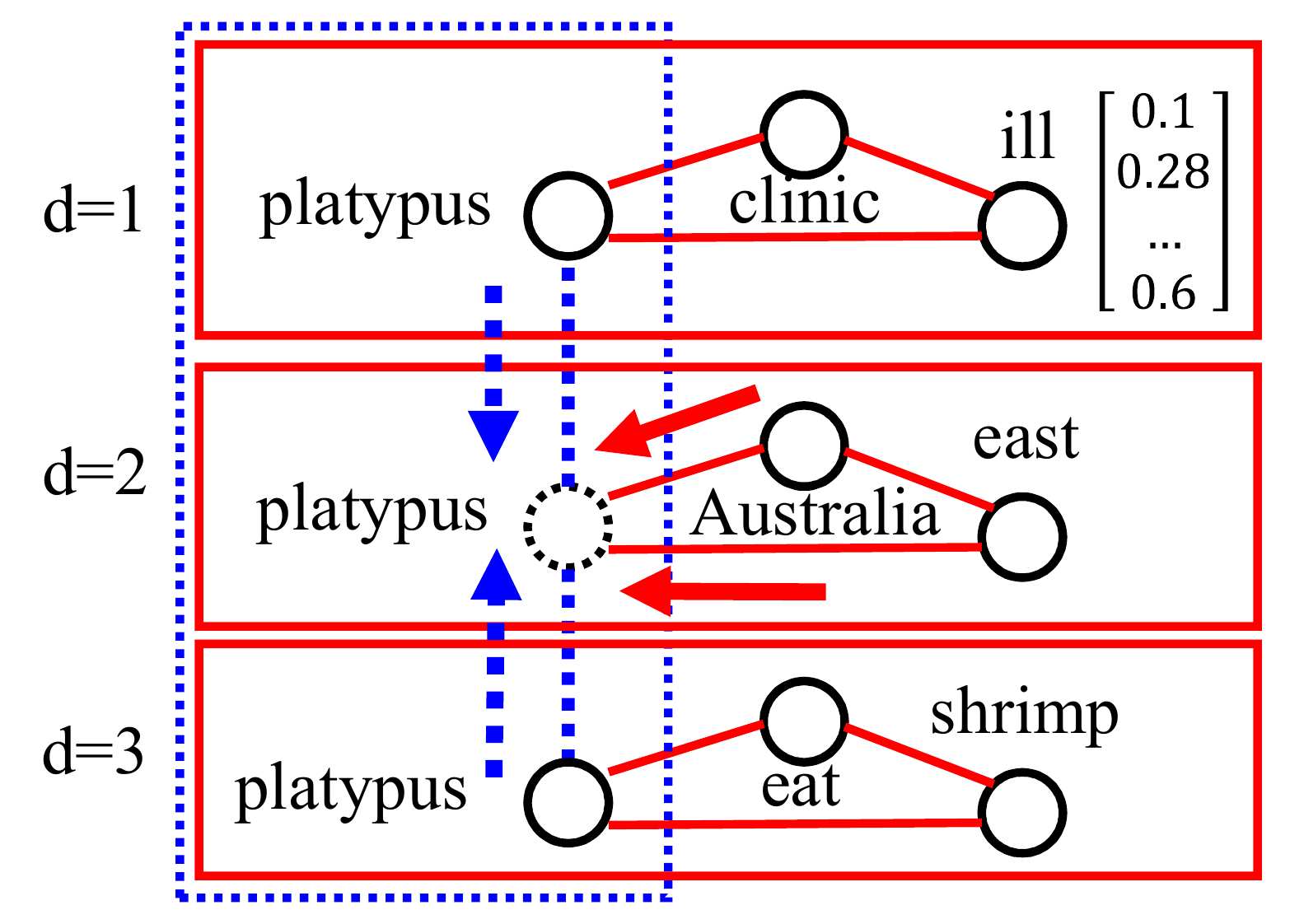}
		\captionof{figure}{Schematic of our network view \\ of LDA}
		\label{fig:networkLDA}
	\end{minipage}%
	\begin{minipage}{.53\textwidth}
		\centering
		\includegraphics[width=.65\linewidth]{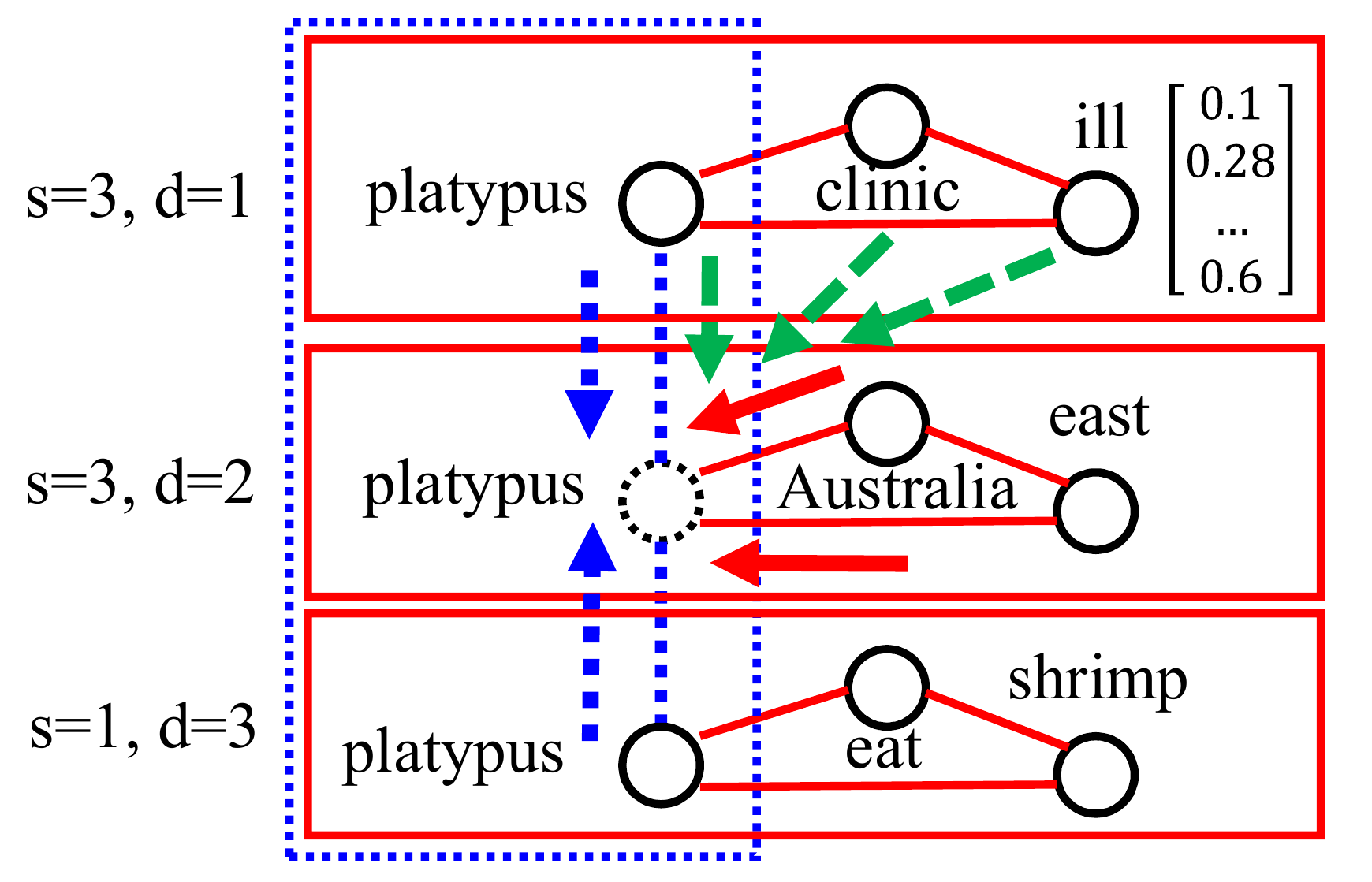}
		\captionof{figure}{Schematic of our network view \\ of supervised LDA}
		\label{fig:networkSLDA}
	\end{minipage}
\end{figure*}

\section{Model}
\subsection{LDA and Neighborhood Aggregation}

The message-passing equation (Eq. (\ref{MesPasLDA})) of LDA can be seen as taking an elementwise product of two neighborhood aggregation operations and normalizing the result for probabilistic interpretation.  We clarify this point with an illustrative example.  Figure \ref{fig:networkLDA} shows a hypothetical corpus with three documents ``platypus, clinic, ill'', ``platypus, Australia, east'' and ``platypus, eat, shrimp''.  Each document--word pair in the corpus has an associated message like the upper-right vector depicted next to the word ``ill'' in document 1.  The red bold lines indicate a relationship describing co-occurring words in a document while the blue dotted lines indicate a relationship describing the same word in the corpus.

Suppose that we want to update the message for the word ``platypus'' found in document 2.  According to the message passing equation (Eq. (\ref{MesPasLDA})), this message is updated by the elementwise product of messages gathered from edges describing co-occurring words in document 2 (as denoted by the red bold arrows) and messages gathered from edges describing the same word in the corpus (as denoted by the blue dotted arrows).  In fact, assuming that the aggregation function in a document is 

\begin{equation}
\label{AggD}
\begin{split}
Agg_{d}^{k}(\mu_{j}(t);\forall j \in Nei_{d}(d,w)) = \\
\Sigma_{w'=1,w' \neq w}^{W} x_{d,w'} \mu_{d,w'}^{k} + \alpha,
\end{split}
\end{equation}

\noindent where $Nei_{d}(d,w)$ denotes all other words in the same document, there are no additional node features to combine and, using $Agg_{d}$ to denote $[Agg_{d}^{1} \cdots Agg_{d}^{K}]$, the neighborhood aggregation operation from co-occurring words in a document can be written as 

\begin{equation}
\begin{split}
NA_{d}(t) = Norm_{K}(Agg_{d}(\mu_{j}(t);\forall j \in Nei_{d}(d,w))),
\end{split}
\end{equation}

\noindent where $Norm_{K}$ is defined to be a normalization function dividing by the sum over topics $1 \leq k \leq K$.  

Following similar reasoning, it is assumed that the aggregation function for the edges describing the same word in the corpus is

\begin{equation}
\label{AggW}
\begin{split}
Agg_{w}^{k}(\mu_{j}(t);\forall j \in Nei_{w}(d,w)) = \\
\Sigma_{d'=1, d' \neq d}^{D} x_{d',w} \mu_{d',w}^{k} + \beta,
\end{split}
\end{equation}

\noindent where $Nei_{w}(d,w)$ denotes all document--word pairs in the corpus that is the same word as $w$ (i.e. nodes in the blue dashed square shown in Figure \ref{fig:networkLDA}) and there are no additional node features to combine (such as pre-trained word embedding vectors described below).  The neighborhood aggregation operation for this neighborhood system can be summarized as

\begin{equation}
\begin{split}
NA_{w}(t) = Norm_{W}(Agg_{w}(\mu_{j}(t);\forall j \in Nei_{w}(d,w))),
\end{split}
\end{equation}

\noindent where $Norm_{W}$ is defined as a normalization function dividing by the sum over words $1 \leq w \leq W$.

The message update equation for a document--word pair $(d,w)$ can thus be summarized by 

\begin{equation}
\begin{split}
\mu_{d,w}(t+1) = Norm_{K}(NA_{d}(t) \odot NA_{w}(t)),
\end{split}
\end{equation}

\noindent where $\odot$ denotes elementwise multiplication.  The messages are normalized in the final step so that they can be used as a proper probability distribution.  

This explanation shows that the message passing equation (Eq. (\ref{MesPasLDA})) of LDA can be seen as an element--wise product of two neighborhood aggregation operations with an additional normalization step for probabilistic interpretation. 

\subsection{Supervised LDA and Neighborhood Aggregation}

We extend the above formulation to incorporate supervisory signals.  Supervisory signals such as label information can be thought of as defining an additional neighborhood system in the above formulation.  We therefore add edges reflecting this additional neighborhood system.  For example, in a classification problem where we have label information for each document, we add edges among document--word pairs that belong to other documents with the same label.  Figure \ref{fig:networkSLDA} illustrates the same hypothetical corpus as shown in Figure \ref{fig:networkLDA}.  The green dashed arrows are the new edges added by the supervisory signal.

We define the aggregation function for this neighborhood system as

\begin{equation}
\label{AggS}
\begin{split}
Agg_{s}^{k}(\mu_{j}(t);\forall j \in Nei_{s}(d,w)) = \\ \Sigma_{d'=1,d' \neq d}^{D} \Sigma_{w=1}^{W} x_{d',w} \mu_{d',w}^{k} 1_{l(d')=l(d)},
\end{split}
\end{equation}

\noindent where the indicator function $1_{l(d')=l(d)}$ is used to select documents with the same label and $l$ is a function that maps a document index $d$ to its corresponding label $s$.  Neighborhood systems can easily be extended to the regression case where documents do not necessarily have exactly the same output value (e.g., real numbers). In this case, we replace the indicator function as $1_{|l(d')-l(d)| < \epsilon}$
for a given $\epsilon > 0$.

From this formulation, the neighborhood aggregation operation from the supervisory signal can be written as 

\begin{equation}
\label{NAS}
\begin{split}
NA_{s}(t) = Norm_{K}(W_{s} Agg_{s}(\mu_{j}(t);\forall Nei_{s}(d,w))),
\end{split}
\end{equation}

\noindent where $Nei_{s}(d,w)$ denotes the neighborhood system of $s$ for document--word pair $(d,w)$ and $W_{s}$ denotes a diagonal matrix with positive entries only.  We train $W_{s}$ using a supervisory signal as is done in neighborhood aggregation algorithms \cite{Hamilton2017}.  

In the standard sum-product algorithm, we take the product of all messages from factors to variables.  However as was noted in \cite{Zeng2013}, a product operation cannot control the messages from different sources (i.e. supervised part and unsupervised part) and we, therefore, take the weighted sum of the two neighborhood aggregation operations to rebalance the two effects.  The whole message updating equation can be summarized as

\begin{equation}
\label{MesNA}
\begin{split}
\mu_{d,w}(t+1) = \\
Norm_{K} (\eta NA_{s}(t) + (1-\eta) NA_{d}(t)) \odot NA_{w}(t)),
\end{split}
\end{equation}

\noindent where $\eta $ controls the strength of the supervisory signal in inferencing the topic label configuration.  This type of rebalancing of the effect from the supervised part and unsupervised is also taken in the traditional supervised LDA approaches where it is well known that the effect of the supervised part is reduced for documents with more words in the standard formulation \cite{Katharopoulos2016}.  Hence, Eq. (\ref{MesNA}) is quite a standard technique, contrary to what it might appear at first glance.

\subsection{Pretrained Word Embedding Vectors}

It is natural to assume that there is an association between pre-trained word embedding vectors \cite{Mikolov2013} and topics, so long as topic models are used to summarize the semantic information in a document.  Hence, learning the association between pre-trained word embedding vectors and topics is important especially at the time of testing when there might be many unseen words in the held-out documents.  If these unobserved words are included in the pre-trained word embedding dictionary and we have trained a function mapping the word embedding vectors to the topics, we can leverage the pre-trained word embedding vectors to predict the topic label configuration more accurately.  This issue was addressed in several recent papers \cite{Das2015}.


Within the neighborhood aggregation framework, the pre-trained word embedding vectors can be modeled as a node attribute vector $v_{i}$ in the notation of Eq. (\ref{NeighborAgg}).  We define the word embedding in the topic distribution transformation as

\begin{equation}
\begin{split}
u_{g(d,w)} = Norm_{K}( \sigma(W_{C}v_{g(d,w)})),
\end{split}
\end{equation}

\noindent where $g$ is a function that maps the document--word index to the word index in the pre-trained word embedding vector dictionary, $W_{C}$ is the $K \times E$ weight matrix transforming word embedding vectors (which we assume to have dimension $E$) to the topic and $\sigma$ is an element--wise nonlinear transformation.

\begin{table*}
	\centering
	\begin{tabular}{ccccccc}
		\toprule
		Dataset & EWS  & & Amazon review & & Subjectivity &  \\
		Performance measure & Cross entropy  & Accuracy & Cross entropy & Accuracy & Cross--entropy
		& Accuracy  \\
		\midrule
		\texttt{SLDA} & 1.541 & 0.336 & 2.398 & 0.415 & 0.661 & 0.650 \\
		\texttt{MedLDA}& - & 0.465 & - & 0.485 & - & 0.821 \\
		\texttt{WE-MLP} & 1.379 & 0.453 & 1.575 & 0.343 & 0.748 & 0.656 \\
		\texttt{LFLDA} & 1.196 & 0.468 & \textbf{1.222} & 0.457 & 0.461 & 0.799 \\
		\texttt{LFDMM} & 1.327 & 0.499 & 1.299 & 0.477 & 0.571 & 0.825 \\
		\texttt{NA-WE-SLDA} & \textbf{1.194} & \textbf{0.504} & 1.304 & \textbf{0.510} & \textbf{0.402} & \textbf{0.829} \\
		\bottomrule
	\end{tabular}
	\caption{Performance of held-out document classification.  EWS stands for Economic Watcher Survey.}
	\label{tab:result:quantitative}
\end{table*}

Using the transformed vector $u_{g}(d,w)$, we define the combine function as

\begin{equation}
\begin{split}
Comb(u_g({d,w}),NA_{w}(t),NA_{d}(t),NA_{s}(t)) = \\
u_{g(d,w)} \odot (\eta NA_{s}(t) + (1-\eta) NA_{w}(t)) \odot NA_{d}(t).
\end{split}
\end{equation}

\noindent The entire message updating equation is now summarized as

\begin{equation}
\label{MesNAWE}
\begin{split}
\mu_{d,w}(t+1) = \\
Norm_{K}(Comb(u_g({d,w}),NA_{w}(t),NA_{d}(t),NA_{s}(t))).
\end{split}
\end{equation}

\noindent This update equation is a particular instance of Eq.~(\ref{NeighborAgg}) assuming identity mapping for $\sigma$ and applying elementwise multiplication for the combine function.

\subsection{Training}

Inference of the messages (i.e., $\mu_{d,w}^{k}$) and estimation of the parameters (i.e., $W_{S}$, $W_{C}$, $S_{A}$, $S_{B}$, $S_{C}$, $T_{A}$, $T_{B}$ and $T_{C}$) can be performed by alternately inferring the topic label configurations given parameters of the model and updating the parameters minimizing the supervised loss given a topic label configuration.  In each iteration, we first infer the topic label configuration for the randomly sampled document--word pairs, then update all the parameters by creating a mini--batch of the document--word pairs and update them using standard stochastic gradient descent technique.  The number of the document--word pairs are set to 5000.  For the regularization parameter governing $W_{S}$ and $W_{C}$, we set it to $0.001$, and for the output function, we set a dropout probability of $0.5$ for regularization.  We also set $\eta$ in our model to be $0.2$ for the economic watcher survey and $0.05$ for the rest, and set the number of hidden units in Eq.(\ref{softmax}) to be $H_{1} = 50$ and $H_{2} = 50$.  We repeat this iteration until convergence. We could also use sparse message passing in the spirit of \cite{Pal2006} to enhance the coherence of the topics, but we leave this for future works.

\begin{table}
	\centering
	\begin{tabular}{ccccccc}
		\toprule
		& EWS  & Amazon review & Subjectivity  \\
		\midrule
		\texttt{LDA} & 0.386 & 0.318 & 0.334 \\
		\texttt{SLDA} & 0.367 & 0.332 & 0.339 \\
		\texttt{MedLDA } & \textbf{0.389} & 0.327 & 0.338 \\
		\texttt{LFLDA}  & 0.377 & \textbf{0.338} & \textbf{0.358} \\
		\texttt{LFDMM}  & 0.363 & 0.323 & 0.323 \\
		\texttt{NA-WE-SLDA} & 0.374 & 0.300 & 0.320 \\
		\bottomrule
	\end{tabular}
	\caption{Topic coherence measurement using the $C_{V}$  measure proposed in [Roder et al., 2015]. EWS stands for Economic Watcher Survey.}
	\label{tab:coherence}
\end{table}

\section{Experiments}

\subsection{Datasets}
We conduct experiments showing the validity of our approach.  We use three datasets in our experiments.  Two of our datasets are data of multiclass-label prediction tasks focusing on sentiment (economic assessments and product reviews) while the other dataset is data of a binary label prediction task focusing on subjectivity.  We summarize the datasets below.

\begin{itemize}
	\item The economic watcher survey, in the table abbreviated as EWS, is a dataset provided by the Cabinet Office of Japan \footnote{The whole dataset is available at \url{http://www5.cao.go.jp/keizai3/watcher_index.html}}.  The purpose of the survey is to gather economic assessments from working people in Japan to comprehend region-by-region economic trends in near real time.  The survey is conducted every month, and each record in the dataset consists of a multi-class label spanning from 1 (i.e., the economy will get worse) to 5 (i.e., the economy will get better) and text describing the reasons behind a judgment.  We use a subset of this dataset describing the assessment of future economic conditions.  We use the 3000 most frequently used words in the corpus and further restrict our attention to records that have more than a total of 20 words in the pre-trained word embedding dictionary to perform a fair comparison among the models\footnote{We use standard morphological analysis software \cite{Kudo2005} to annotate each word.}.   We randomly sample 5000 records for training, development, and testing.  For the pre-trained word embedding vectors describing Japanese words, we use word2vec vectors provided by \cite{Suzuki2016}\footnote{These are available for download at \url{http://www.cl.ecei.tohoku.ac.jp/~m-suzuki/jawiki_vector/}}.
	\item Amazon review data are a dataset of gathered ratings and review information \cite{McAuley2015}.  We use a subset of the five-core apps for android dataset\footnote{The term five-core implies a subset of the data in which there are at least five reviews for all items and reviewers.} and use the 3000 most frequently used words in the corpus.  We randomly sample 5000 records for training, development, and testing focusing on reviews that have more than 20 words excluding stop words in the pre-trained word embedding dictionary\footnote{The whole dataset is available at \url{http://jmcauley.ucsd.edu/data/amazon/}.}.   For the pre-trained word embedding vectors describing English words, we use the word2vec embedding vectors provided by Google \cite{Mikolov2013}.
	\item Subjectivity data are a dataset provided by \cite{PangLee2004}, who gathered subjective (label 0) and objective (label 1) snippets from Rotten Tomatoes and the Internet Movie Database.  We focus on snippets that have more than nine words and sample 1000 snippets each for training, development and testing\footnote{The whole dataset is available at \url{http://ws.cs.cornell.edu/people/pabo/movie-review-data}.}.  Other settings are the same as those of the Amazon review data.
\end{itemize}

\subsection{Classification Performance}

The main goal in this section is to compare the classification performance of our proposed models with that of the state-of-the-art supervised LDA implementations (which we denote as SLDA\cite{Katharopoulos2016} and MedLDA\cite{Zhu2009}), nonlinear prediction using pre-trained word embedding vectors only (which we denote as WE-MLP) and topic models incorporating pre-trained word embedding vectors (which we denote as LFLDA\cite{Nguyen2015}  and LFDMM\cite{Nguyen2015}.  The difference between LFLDA and LFDMM is that LFDMM assumes that all the words in a document share the same topic.  We use the default settings defined in the codes provided by the authors.  We also fix the number of topics to 20 for all experiments performed in this section.  We compare the performance of these models which we denote as NA-WE-SLDA. 

For WE-MLP, we use the average of all the pre-trained word embedding vectors found in a document as the feature describing a document and we connect this feature to our supervisory signal using a simple multilayer perceptron with softmax output.  For LFLDA and LFDMM we use the document topic distributions as the feature describing a document and also connect this feature to our supervisory signal using a simple multilayer perceptron with softmax output.  Parameters (e.g., the number of hidden units) of these models was found by utilizing the development dataset.

Table~\ref{tab:result:quantitative} summarizes the results.  Cross-entropy denotes the cross-entropy loss and accuracy denotes the proportion of correct predictions in the held-out documents.  For MedLDA we only report accuracy because the method does not use the softmax output function.  We see that NA-WE-SLDA is the best performing model in all cases and sometimes beats SLDA and MedLDA substantially, especially for multi-label classification tasks when measuring with accuracy.  

\subsection{Comparison with Deep Learning}

Here we compare our model with the state-of-the-art deep learning method \cite{Kim2014} varying the number of documents from 1000 to 5000.  We chose the filter sizes, number of filters and dropout probability by using development data set.  We see that for smaller document size our method performs better than the deep learning method, but as the number of the document increases the power of deep learning catches up, and our model is out-beaten as expected.  However, our performance for a smaller number is worth noting.

\begin{table}
	\centering
	\resizebox{0.47\textwidth}{!}{
	\begin{tabular}{lcccc}	
		\toprule
		No. Docs & CNN & & NA-WE-SLDA &   \\
		 & Cross entropy  & Accuracy & Cross entropy & Accuracy  \\
		\midrule
		1000 & 1.265 & 0.483 & \textbf{1.212} & \textbf{0.498} \\
		2000 & \textbf{1.192} & 0.502 & 1.217 & \textbf{0.508} \\
		3000 &  \textbf{1.114} &  \textbf{0.546} & 1.221 & 0.505 \\
		4000 &  \textbf{1.092} &  \textbf{0.556} & 1.221 & 0.513 \\
		5000 &  \textbf{1.099} &  \textbf{0.553} & 1.194 & 0.504 \\
		\bottomrule
	\end{tabular}}
	\caption{Comparison with deep learning method for the Economic Watcher Survey.}
	\label{tab:result}
\end{table}

\subsection{Topic Coherence}

Our goal in this section is to compare topic coherence among the supervised topic models that we evaluated in the previous section and the plain LDA model \cite{Blei2003}.  The purpose of this experiment is that adding complex structures to topic models might have a side effect of downgrading interpretability \cite{ChangReadTea}.   We want to examine whether our models sacrifice interpretability to achieve better predictive performance.  

Traditionally, topic coherence is evaluated on the basis of the perplexity measure, which is basically a measure of how likely a model predicts words in the held-out documents.  However, as was shown in \cite{ChangReadTea}, higher perplexity does not necessarily correlate to better human interpretability and various topic coherence measures have been proposed to address this issue \cite{TopicEval2015}.  The present paper uses the $C_{V}$ measure proposed in \cite{TopicEval2015}, which is a coherence measure that combines existing basic coherence measures.   It was shown that the $C_{V}$ measure is the best coherence measure among 237,912 coherence measures compared in their paper \cite{TopicEval2015}.  We employ the source code provided by the authors\footnote{The code is available at \url{https://github.com/dice-group/Palmetto}} to calculate the coherence measure in our experiment focusing on the top 15 words in the topic distribution\footnote{For the economy watcher survey, we translated words from Japanese to English.}.

Table~\ref{tab:coherence} summarizes the results.  We see that our method has a price to pay for its better predictive accuracy.  For all the data set, our model shows inferior coherence compared to other methods.  Our second observation is that except for the economic watcher survey dataset and our model, models with word embedding vectors slightly outperforms the model without word embedding vectors.  Our model even utilizing word embedding vectors, could not compensate for the loss made from the nonlinear output function and lack of sparsification, but still provides comparable performance.

To gain further insights into our learned topics, in Table~\ref{tab:qualitative1}, Table~\ref{tab:qualitative2} and Table~\ref{tab:qualitative3}, we report the top six words of the 20 topics learned by LDA, SLDA, and NA-WE-SLDA respectively for the economic watcher survey dataset\footnote{Stop words are omitted.}.  First of all we see several topics that does not appear in LDA such as ``ecology, payment, automobile, subsidy, sale, car'' (i.e., Topic 1) in SLDA and ``ecology, subsidy, car, payment, last, year'' (i.e., Topic 8) in NA-WE-SLDA highlighting the effect of the supervisory signals.  It is also worth mentioning that although the topic corresponding to the great east Japan earthquake does seem to appear in LDA as in Topic 7 (i.e., ``customer, last, year, situation, disaster, east''), the topic corresponding to the earthquake event is easier to spot in SLDA and NA-WE-SLDA, where the former focuses more on the impact on the car industry (i.e., Topic 13) and the latter focuses more on the nuclear incident (i.e., Topic 9).  The difference between SLDA and NA-WE-SLDA could be seen in topics that focus on temporary dispatch workers\footnote{Temporary workers sent by staffing agencies.}.  While Topic 15 in SLDA seems to focus on dispatch workers in small and medium-sized companies, Topic 16 in NA-WE-SLDA focuses more on the end of the fiscal year which corresponds to the timing when dispatch workers are more susceptible to layoffs.

\section{Conclusion}

We showed that topic modeling with word embedding can be viewed as implementing a neighborhood aggregation algorithm where the messages are passed through a network defined over words.  By exploiting the network view of topic models, we proposed new ways to model and infer supervised topic models equipped with a nonlinear output function.  Our extensive experiments performed over a range of datasets showed the validity of our approach.

\begin{table}
	\centering
	\resizebox{0.47\textwidth}{!}{
		\begin{tabular}{ccccccccccccccccccccc}
			\toprule
			Topic number  & Top 6 words  \\
			\midrule
			Topic 1 & company, economy, like, gain, production, job \\
			Topic 2 & person, job, hunting, situation, company, increase \\
			Topic 3 & customer, good, goods, situation, sales, continue \\
			Topic 4 & economy, company, expect, feel, continue, many \\
			Topic 5 & consumption, tax, increase, goods, customer, life \\
			Topic 6 & price, product, increase, unit, situation, impact \\
			Topic 7 & customer, last, year, situation, disaster, east \\
			Topic 8 & good, economy, months, number, company, customer  \\
			Topic 9 & consumption, tax, increase, last, minute, demand \\
			Topic 10 & number, last, year, sales, visitor, future \\
			Topic 11 & customer, store, sales, increase, tourism, people \\ 
			Topic 12 & economy, company, yen, weaker, consumption, stronger \\
			Topic 13 & last, year, travel, percent, ratio, reservation \\
			Topic 14 & economy, look, demand, person, number, like \\
			Topic 15 & job, offer, number, last, year, tendency \\
			Topic 16 & industry, production, impact, job, offer, none \\ 
			Topic 17 & consumption, last, year, situation, ecology, impact \\ 
			Topic 18 & consumption, situation, customer, economy, continue, travel \\
			Topic 19 & months, 3, 2, good, situation, ahead \\
			Topic 20 & consumption, economy, company, recover, business, gain \\
			\bottomrule
	\end{tabular}}
	\caption{Topics learned by LDA.}
	\label{tab:qualitative1}
\end{table}

\begin{table}
	\centering
	\resizebox{0.47\textwidth}{!}{
		\begin{tabular}{ccccccccccccccccccccc}
			\toprule
			Topic number  & Top 6 words  \\
			\midrule
			Topic 1 & ecology, payment, automobile, subsidy, sale, car \\
			Topic 2 & expect, increase, possible, sales, movement, target \\
			Topic 3 & consumption, tax, increase, demand, last, minute \\
			Topic 4 & good, customer, gain, economy, expect, little  \\
			Topic 5 & industry, order, production, job, fiscal, year \\
			Topic 6 & Tokyo, city, sign, good, Olympic, go \\
			Topic 7 & goods, sales, product, unit, price, customer \\ 
			Topic 8 & good, economy, none, situation, change, customer  \\
			Topic 9 & price, fee, store, soaring, continue, increase \\
			Topic 10 & yen, price, stronger, economy, impact, weaker \\
			Topic 11 & people, tourism, customer, foreign, use, expect \\
			Topic 12 & economy, customer, bad, consumption, situation, like \\
			Topic 13 & east, Japan, disaster, impact, car, commerce \\
			Topic 14 & reservation, travel, last, year, situation, customer \\
			Topic 15 & company, hire, small, medium, sized, dipatch \\
			Topic 16 & consumption, tax, increase, demand, last, minute \\
			Topic 17 & month, last, year, 10, 12, sales \\
			Topic 18 & months, 3, 2, ahead, number, situation \\
			Topic 19 & yen, economy, consumption, stronger, weaker, impact \\
			Topic 20 & job, offer, number, last, year, percent \\
			\bottomrule
	\end{tabular}}
	\caption{Topics learned by SLDA.}
	\label{tab:qualitative2}
\end{table}

\begin{table}
	\centering	
	\resizebox{0.47\textwidth}{!}{
		\begin{tabular}{ccccccccccccccccccccc}
			\toprule
			Topic number  & Top 6 words  \\
			\midrule
			Topic 1 & situation, economy, number, future, good, change \\
			Topic 2 & good, USA, change, expect, continue, economy \\
			Topic 3 & consumption, good, economy, customer, situation, expect \\
			Topic 4 & economy, situation, consumption, customer, company, person \\
			Topic 5 & last, year, good, number, month, customer \\
			Topic 6 & customer, good, economy, future, consumption, situation \\
			Topic 7 & consumption, customer, situation, economy, good, none \\
			Topic 8 & ecology, subsidy, car, payment, last, year  \\
			Topic 9 & nuclear, Fukushima, one, place, power, generation \\
			Topic 10 & last, year, 2, ratio, 3, number \\
			Topic 11 & consumption, situation, customer, economy, company, continue \\
			Topic 12 & job, offer, last, year, months, number \\
			Topic 13 & consumption, customer, situation, impact, economy, continue \\
			Topic 14 & consumption, tax, increase, last, year, situation \\
			Topic 15 & industry, production, company, continue, economy, impact \\
			Topic 16 & fiscal, year, dispatch, worker, person, end \\
			Topic 17 & consumption, good, economy, customer, expect, company \\
			Topic 18 & consumption, tax, increase, demand, last, minute \\
			Topic 19 & last, year, ratio, month, percent, number \\
			Topic 20 & economy, consumption, situation, company, impact, continue \\
			\bottomrule
	\end{tabular}}
	\caption{Topics learned by NA-WE-SLDA.}
	\label{tab:qualitative3}
\end{table}

\section*{Acknowledgements}
This work was supported by JSPS KAKENHI Grant Number JP17K17663.  We thank Glenn Pennycook, MSc, from Edanz Group (www.edanzediting.com/ac) for editing a draft of this manuscript.

\newpage

\bibliographystyle{named}
\bibliography{HISANO_MANU_FIN}

\end{document}